# A novel and automatic pectoral muscle identification algorithm for mediolateral oblique (MLO) view mammograms using ImageJ

Chao Wang

Wolfson Institute of Preventive Medicine
Queen Mary University of London
Charterhouse Square
London
United Kingdom
EC1M 6BQ
Tel: +442078823541; Email: chao.wang@qmul.ac.uk

## Abstract

Pectoral muscle identification is often required for breast cancer risk analysis, such as estimating breast density. Traditional methods are overwhelmingly based on manual visual assessment or straight line fitting for the pectoral muscle boundary, which are inefficient and inaccurate since pectoral muscle in mammograms can have curved boundaries.

This paper proposes a novel and automatic pectoral muscle identification algorithm for MLO view mammograms. It is suitable for both scanned film and full field digital mammograms. This algorithm is demonstrated using a public domain software ImageJ. A validation of this algorithm has been performed using real-world data and it shows promising result.

## 1   Introduction

Mammograms can be useful for diagnosing breast cancer. For example, mammographic density is a well-established biomarker for breast cancer (Assi, Warwick, Cuzick, & Duffy, 2012). Breast mainly contains fibroglandular tissue and fat, and mammographic density is the region of tissues that is radiopaque and appears white on mammograms. One popular quantification of breast density is the percentage of dense area within the total breast area, often referred to as "percent density" (PD) which has attracted a lot of research interests. For example, it is estimated that 16% of all breast cancers and 26% of breast cancers in women aged under 56 years were attributable to breast densities over 50% (Boyd et al., 2006). In order to estimate PD, it is required to accurately identify both regions of breast itself and the dense areas within the breast boundary. Mammograms usually have two views, namely craniocaudal (CC) view and mediolateral oblique (MLO) view. The former is a "top-to-bottom" view while the latter is a side view. The issue with a MLO view mammogram is that it usually contains a large area of pectoral muscle in the image, which creates a problem of identifying the boundary of a breast. This further makes it difficult to accurately estimate the PD and ultimately cancer risk.

In addition to the problem in the application of estimating PD, a recent study has found that the image feature (i.e. the mean pixel intensity value) of pectoral muscle itself in a mammogram may be an independent risk factor for breast cancer (Cheddad et al., 2014). This again, calls for an appropriate method to correctly identify the boundary of pectoral muscle in a MLO view mammogram.

A traditional and straightforward method to identify the pectoral muscle is visual assessment. This requires a human reader to look at the mammogram and manually identify the pectoral muscle region. This is a popular method used in, for instance, Cumulus software (Byng, Boyd, Fishell, Jong, & Yaffe, 1994), a "gold standard" for PD estimation. However, identifying pectoral muscle manually can be labour intensive and time consuming. This is especially a problem in a large scale screening study. There has been a trend advocating fully automated method for mammogram analysis (e.g. (Keller et al., 2012)), and this requires pectoral muscle identification also to be fully automated.

Several authors have proposed automatic pectoral muscle identification method. For instance Karssemeijer (1998) proposed a straight line fitting method with classical Hough transform to separate breast and pectoral muscle. The rationale behind this type of method is that the pectoral muscle is usually located in upper left or right part in the image and the boundary of the pectoral muscle can be represented by a straight line at an appropriate angle. This method has then been used in many studies, such as Nielsen et al. (2011). Indeed, as pointed out by Camilus, Govindan, and Sathidevi (2011), most methods in the literature are based on straight line fitting. However, the boundary of pectoral muscle is not always straight – in fact a curved boundary is not uncommon in MLO view mammograms. Therefore the straight line assumption may result in incorrect separation of breast and pectoral muscle.

Camilus et al. (2011) proposed an alternative method based on watershed transformation which can find the curved boundary. While this method seems to work well, it is yet to see how well it works on other validation datasets. In particular, their method was developed based on scanned film mammograms, so it is interesting to see how well it works on full field digital mammograms which have been increasingly popular.

This paper introduces another algorithm using morphological reconstruction and automatic threshold method. This method can handle both scanned film and digital mammograms, and can find the curved boundary of pectoral muscles. The algorithm was implemented in the popular image processing software ImageJ which is very efficient for processing large images. The next section below details the proposed algorithm, followed by discussion and conclusion.

## 2 The algorithm

The essential idea of this algorithm is to dampen the brightness of the breast area before applying an automatic threshold method. Intuitively, one might consider using an automatic threshold method to identify pectoral muscle since the pectoral muscle is much brighter than the breast. However, in mammograms, some tissues can be very bright so they can be wrongly classified as pectoral muscle. This algorithm addresses this issue by dampen the brightness of the breast area. After threholding, a morphological open operation was performed so as to remove extra wrongly classified portions.

The details of this algorithm are described in each step as follows. For illustration, a digital raw mammogram (i.e. "for processing") was used as an example.

### 2.1 Contrast enhancement

Digital raw mammograms have limited contrast. We enhance the contrast by using a technique named "windowing". The "windowing", as the name suggests, define a value range (i.e. "window") between the original minimum and maximum pixel intensity values in the image. Then this new range of intensity values were used such that the upper bound of the "window" becomes the new maximum value for the image, and similarly, the lower bound of the "window" becomes the new minimum value for the image. The rest of pixel intensities were then linearly stretched to be mapped to new intensity values between the minimum and maximum values. For the illustration of this study, the breast was firstly segmented from the background in an image by thresholding. Then the original minimum value of the breast region were used to be the lower bound of the window; and the value at the 75 percentile of the pixel value range within the breast was used to be the upper bound.

Figure 1 shows the effect of applying windowing operation.

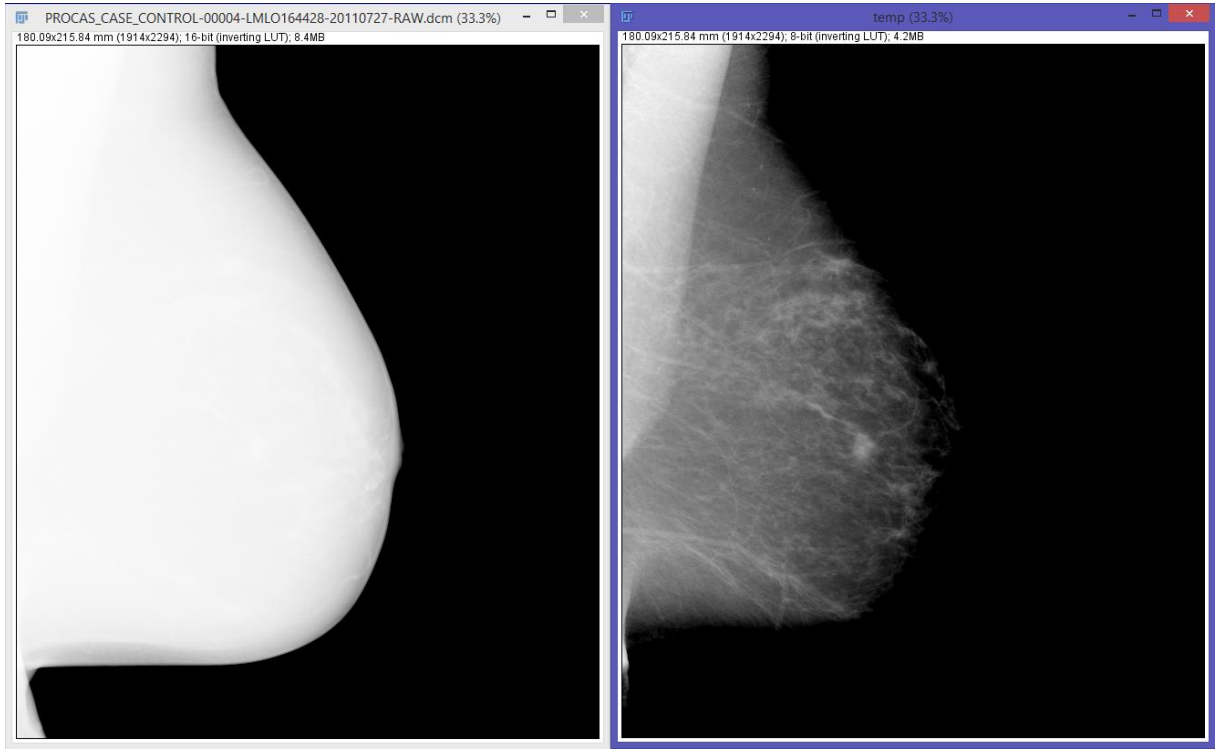

*Figure 1 Windowing (left: digital raw image; right: image after applying windowing operation)*

As can be seen, the boundary of the pectoral muscle is clearly curved.

## 2.2 Morphological reconstruction

Since we know that the pectoral muscle is located at upper position in a mammogram, we can easily create a "marker" that is at either the upper left or right corner, to be used for morphological reconstruction. This has been done by firstly selecting first few lines of an image, and then apply a threshold method using Otsu algorithm. The selected "marker" is shown in Figure 2:

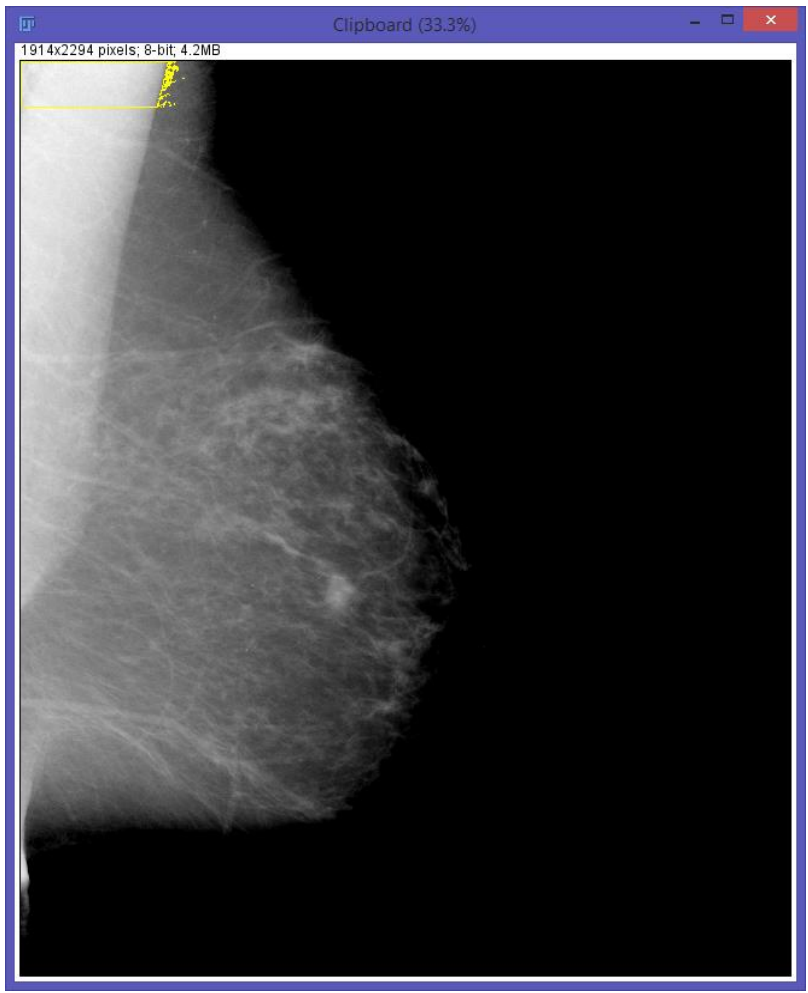

*Figure 2 Selected "marker" in yellow box.*

Having selected the "marker", we next use the windowed image (as shown in Figure 1) as a "mask", and perform the grayscale geodesic reconstruction by dilation using the "MorphoLibJ" ImageJ plug-in (http://github.com/ijpb/MorphoLibJ). The principle of geodesic reconstruction is to perform repeated dilations on the "marker", subject to a "mask" image until no more modification occurs. The resulting image is presented Figure 3. As can be seen, the overall shape of the breast was reconstructed, but the brightness of dense tissues within the breast has been greatly dampened.

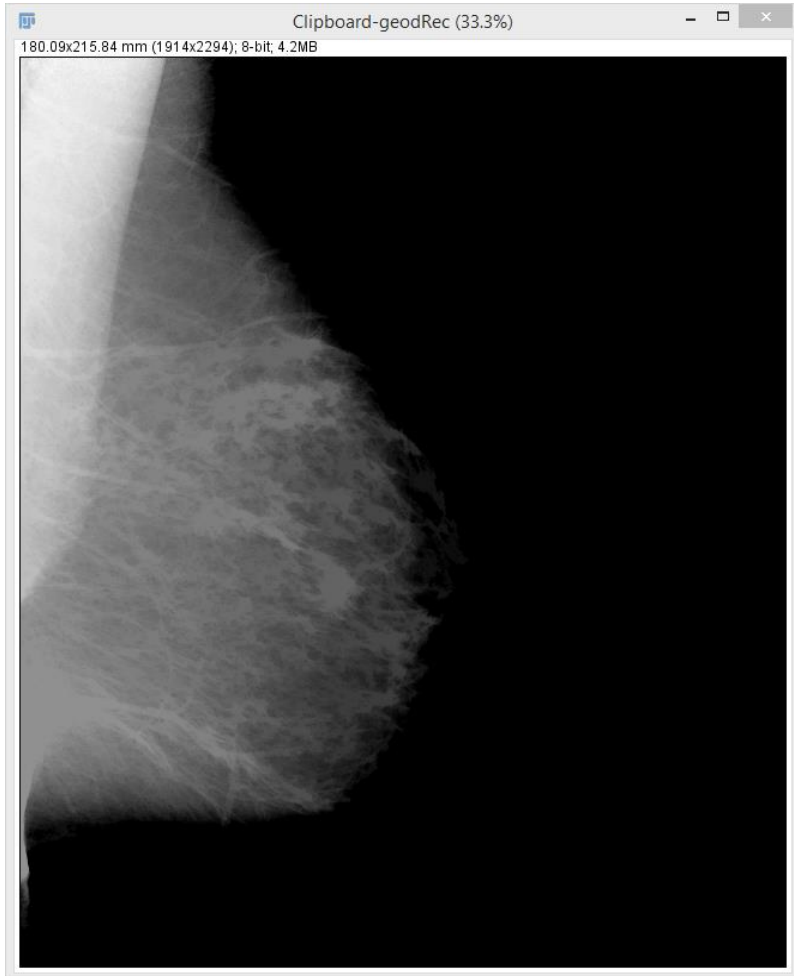

*Figure 3 Mammogram after morphological reconstruction*

## 2.3   Automatic thresholding

Having obtained the reconstructed mammogram, an automatic thresholding can be applied. The maximum entropy thresholding method (Kapur, Sahoo, & Wong, 1985) was adopted since it outperforms other methods after visual checks. The resulting image, with pectoral muscle (i.e. the portion of pixels with higher values after thresholding) highlighted in red colour, is presented in Figure 4. While the resulting image largely identifies the pectoral muscle, it is not very accurate, since it is clear that some dense tissues within the breast are also classified as pectoral muscle. An additional step is required to correct these wrongly classified tissues, as discussed in the next section.

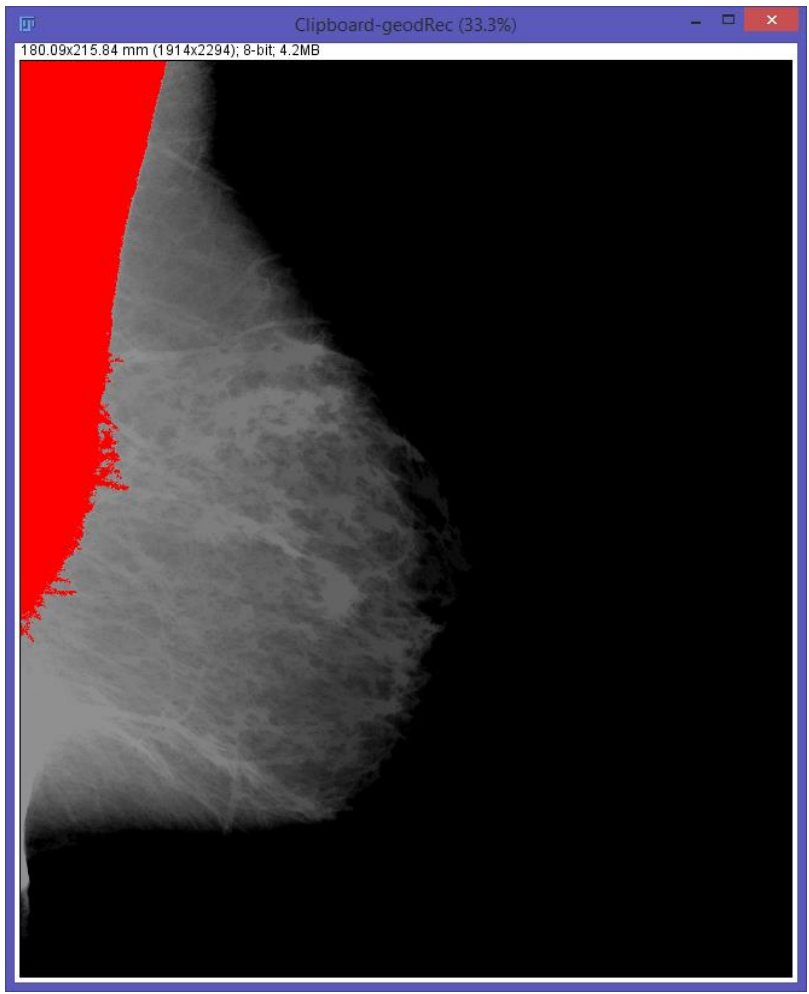

*Figure 4 Maximum entropy thresholding*

## 2.4 Further adjustment and smoothing the muscle boundary

This can be achieved by using a combination of morphological close and open operations. The thresholded image (Figure 4) was firstly converted to a binary (i.e. black & white) image, then a morphological close and open operations were applied. The close operation has the effect of "filling the holes", which can correct the scenario that if an area is part of pectoral muscle but wrongly classified as breast. The open operation has the effect of removing areas that are smaller than the size of the structuring element, which serves the purpose of correcting small bright tissues wrongly classified as pectoral muscle. The structuring elements used in both cases are disk shaped. The effects of the morphological close and open operations are illustrated in Figure 5. Using the resulting image (the right image in Figure 5) as a mask, it is straightforward to find the boundary of the pectoral muscle in the mammogram. Figure 6 shows the final result with the boundary of the pectoral muscle highlighted in yellow curves on the windowed mammogram.

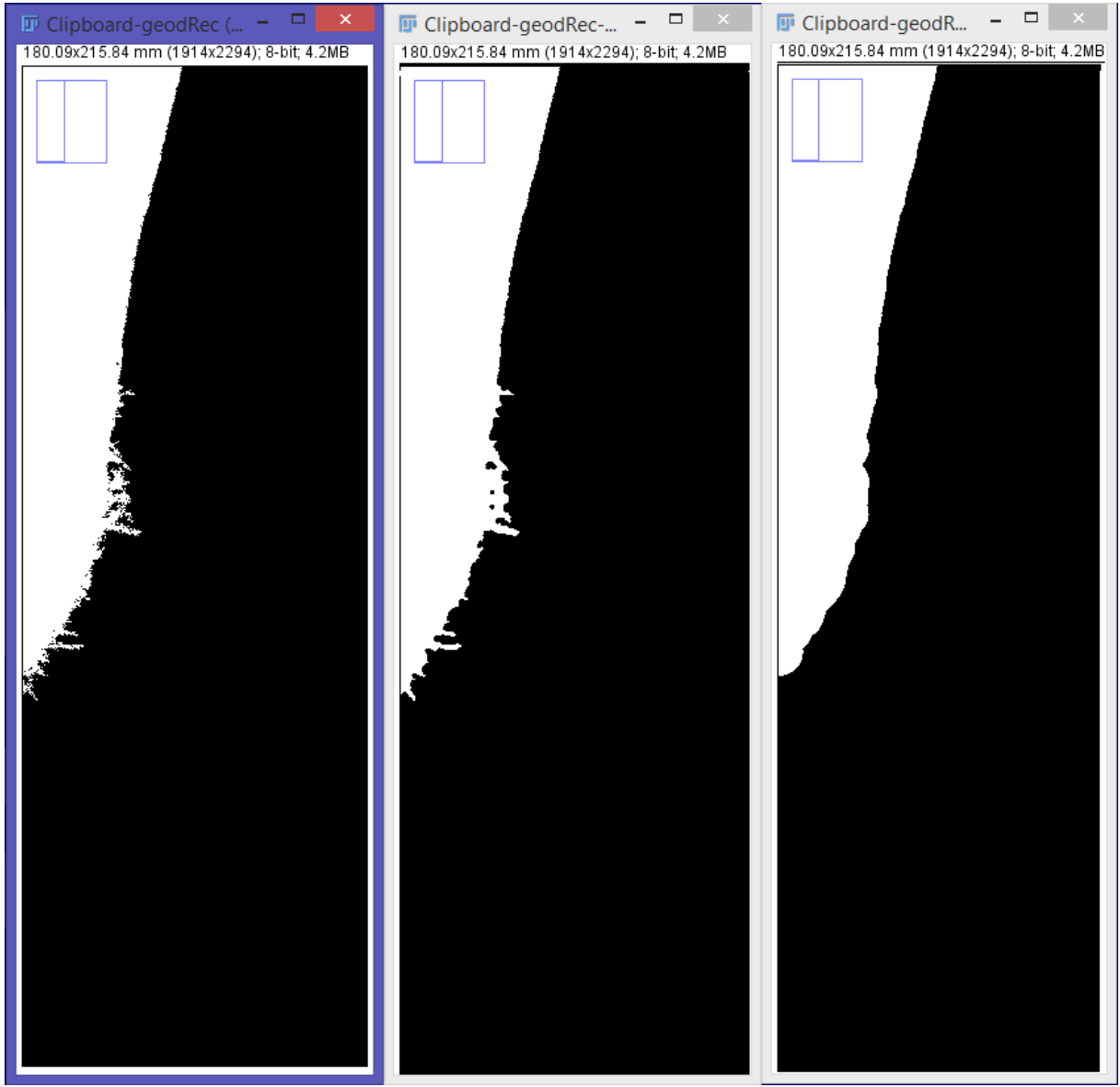

*Figure 5 Demonstration of morphological close and open operations (left: original thresholded image; middle: resulting image after applying close operation; right: resulting image after applying open operation)*

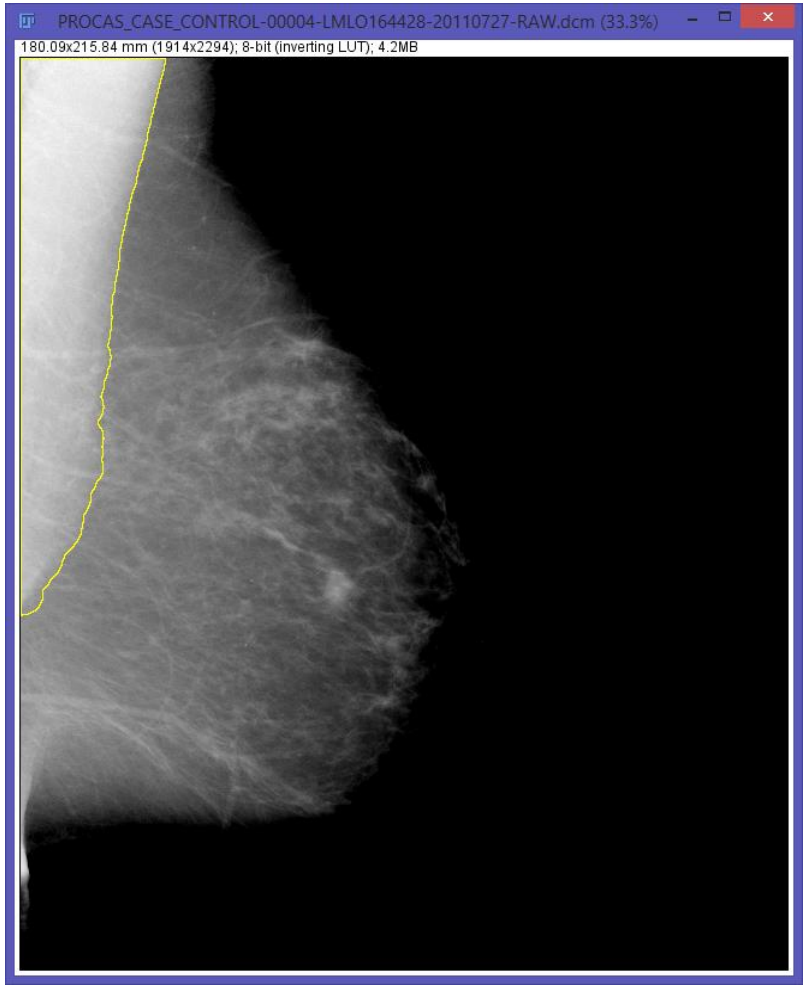

*Figure 6 The final result. The boundary of the pectoral muscle is highlighted in yellow curves.*

## 3   Validation and discussion

To validate this algorithm, mammograms were collected from the Predicting Risk Of breast Cancer At Screening (PROCAS) study, at University Hospital of South Manchester[1]. A total number of 2,564 mammograms were made available including both left and right MLO views. Whether the pectoral muscle was identified correctly was examined by visual assessment done manually. The results show that, among 2,564 mammograms, 104 mammograms were incorrectly identified due to dense tissues wrongly classified as pectoral muscle; 222 were incorrectly identified due to parts of the pectoral muscle wrongly classified as breast. In total, the error rate is 12.71%. In other words, 87.29% of mammograms were identified correctly, which seems to be comparable to Camilus et al. (2011).

The mammograms from the PROCAS study were produced by the GE Medical Systems. It would be interesting to validate this algorithm on mammograms from other manufacturers such as Philips and Sectra. In addition, even with the same machine, the image properties could change significantly due to operators' different settings. It is likely that some minor adjustments to the parameters used in the algorithm may be required to better capture the pectoral muscle boundary in different settings.

---

[1] http://www.uhsm.nhs.uk/research/Pages/PROCASstudy.aspx

## 4  Conclusion

The proposed algorithm in this paper is efficient and can accurately identify the pectoral muscle in a MLO view mammogram.

## 5  Acknowledgement

The author would like to thank the University Hospital of South Manchester for providing mammograms. The PROCAS study is funded by the National Institute for Health Research, and the Genesis Appeal. The research outcome presented in this paper is the result of a research program funded by the Cancer Research UK.

The views expressed in this paper are those of the author alone. The proposed algorithm is implemented in ImageJ, and an example of ImageJ macro script (tested on digital raw mammograms from the GE Medical Systems) can be downloaded at: https://figshare.com/s/48604e512b43e3426c52 (doi: 10.6084/m9.figshare.3081499).